# review articles

DOI:10.1145/3571724

Uniting data-centric perspectives and concepts to trace the foundations of DCAI.

BY MOHAMMAD HOSSEIN JARRAHI, ALI MEMARIANI, AND SHION GUHA

# The Principles of Data-Centric AI

THE ROLE OF data and its quality in supporting AI systems is gaining prominence and giving rise to the concept of data-centric AI (DCAI), which breaks away from widespread model-centric approaches. The flurry of conversation around DCAI can be credited to a recent campaign by Andrew Ng, an AI pioneer, and his colleagues. However, DCAI is a culmination of concerns and efforts around improving data quality in AI projects. DCAI can be understood as an emerging term for a wealth of preceding practices and research work around data quality that complements structured frameworks such as human-centered data science.[4,5] As such, the nature of 'data work' itself is not necessarily new.[35] However, over the years, the actual data work in AI projects comes mostly from individual initiatives, and/or from piecemeal and ad hoc efforts. A lack of attention to data excellence and quality of data has resulted in underwhelming outcomes for AI systems, particularly those deployed in high-stake domains such as medical diagnosis.[35] DCAI magnifies the role of data throughout the AI life cycle and stretches its lifespan beyond the so-called "preprocessing step" in model-centric AI.

The primary goal of *model-centric AI* has been to improve model performance via optimizing the learning parameters and hyper-parameters of a given model. The perceived and measurable success of the model comes from both the algorithm's design and the sophistication of the actual model, and not the data used to construct and validate the model. The prototypical model-centric approach in developing AI systems leaves little opportunity for systematically and progressively revising and improving data quality.[5] Data preparation is performed only at the onset and through "preprocessing" steps in machine learning (ML), establishing a static approach toward data quality. The burden of dealing with data issues (for example, data noise) mostly rests with models (Figure 1 demonstrates how the dataset may remain mostly unchanged throughout the life cycle). This may reflect the prevalent norm of 'data indifference' in the AI community that implicitly relegates the role of data to merely fuel for

» **key insights**

■ DCAI is an emerging paradigm that emphasizes the importance of data quality and dynamism in AI systems, using an iterative, systematic approach.

■ DCAI is redefining the role of data from being merely a preprocessing concern to a continuous improvement factor, encouraging consistent enhancement of both data and model throughout the AI life cycle by incorporating strategies such as data augmentation.

■ A specific contribution of this article is its focus on the human-centered nature of data that feeds AI systems, presenting data as a sociotechnical system, embodying both technological elements and social norms, and biases.

IMAGE BY OLLOMY

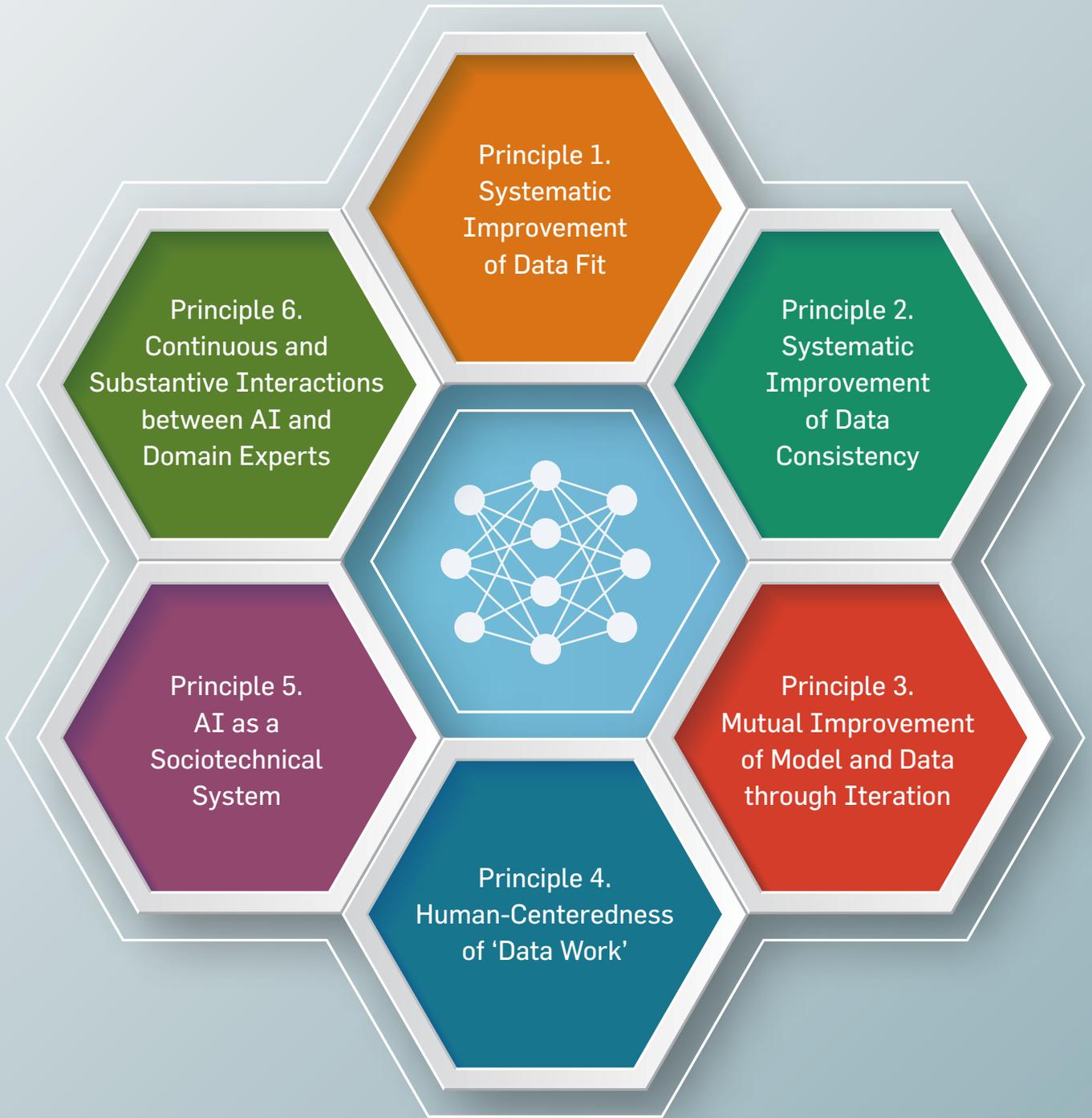

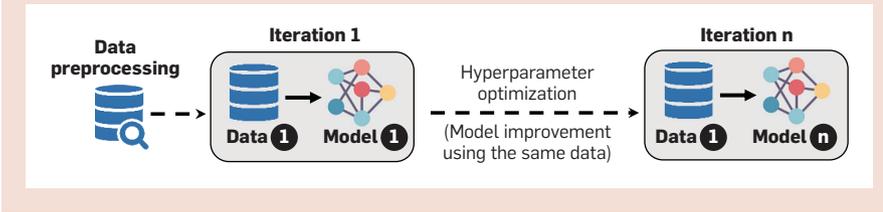

Figure 1. The prototypical life cycle of model-centric AI.

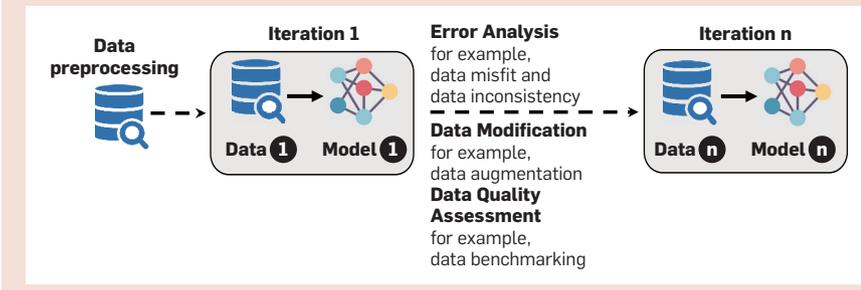

Figure 2. The iterative life cycle of data-centric AI, integrating data improvement.

model training.[20] Data indifference can result in 'data cascades' as compounding events creating negative and unpredictable consequences in downstream AI deployments.[35]

Over the past few years, the democratization of deep learning models through open source communities, together with the rise of paid APIs, has provided unprecedented opportunities for easier access to many optimized models in different domains. Even with this progress, piecemeal quality control of data remains a vexing issue.[27] As a result, fresh attention to data is partly driven by the relative improvement in the models; researchers and developers now might have more bandwidth to tackle data issues. More importantly, the increasing usage of applications of AI in real-world contexts make it clear that data is often "messy" and therefore integration of these datasets without rigorous quality control regimens may result in a "garbage in = garbage out" problem.[5] Attention to data and its context-specificity (rather than just the model) can help generalize AI models for use in a larger number of domains, allowing these models to address more complex real-world problems.

*Data-centric AI.* DCAI advocates for a systematic and iterative approach to dealing with data issues. The foundational premise of DCAI is that AI applications are dynamic, and so are the datasets that train and sustain them. A model that is designed with a DCAI approach recognizes and strives to capture the dynamism of data as an ever-evolving piece of infrastructure in an AI system. Data is the backbone of AI systems across the board,[39] but the application of DCAI is even more vital beyond the frontiers of AI research and practice. We think a DCAI approach is relevant not only to Big Tech companies with access to ultra-large datasets but also to industries and with use cases where training samples are typically smaller (for example, manufacturing) or harder to produce given different types of regulatory or practical constraints (healthcare regulations).

In the DCAI approach, the effectiveness of the AI system is assessed based on both the performance of the model and the quality of the data intertwined with the model. Data and models are both evaluated and refined through numerous iterations (see Figure 2). These iterations can range from initial training to deployment in production. Each iteration (re)trains and refines the model and improves the quality of the data.

After each iteration, three typical steps need to be taken. In the first step, the team conducts *error analysis* to identify sources of error based on *data fit* (how effectively the data helps train the model based on an accurate representation of the real-world problem at hand) and *data consistency* (accuracy and consistency of data annotation).

The second step uses findings from error analysis, looking at the sources of errors in the data, and implements ways to overcome these flaws or opportunities for improvement. For instance, *data augmentation* is a strategy to raise data fit by introducing synthetic data to diversify the data sampling approach.[23]

The third step involves systematic assessment of data quality and can build on *data benchmarking* as a major approach. Data benchmarking broadly refers to strategies to compare the quality of data in training and test sets across two consecutive iterations. Examples of strategies for effective data benchmarking are: unsupervised learning approaches that reveal data misfits such as data imbalance and detection of anomalies, for example, Admad et al.;[1] applying dimensionality reduction to benchmark unstructured data (for example, images and sequences) using an unsupervised algorithm, for example, Becht et al.;[6] data-validation techniques inspired by database systems in deriving data-quality metrics that are tailor-fit for AI systems, for example, Boehm et al.;[8] and finally monitoring the improvements in the performance of the AI model itself, which is the ultimate measure of data quality in developing AI systems.[26]

Here, we provide a formulation of six guiding principles of DCAI by bringing together already existing but dispersed data concepts and practices, especially recent developments in human-centered AI and human-centered data science.[4,38] These principles (particularly principles 4 and 5) recognize that data is a sociotechnical construct: created, manipulated, and interpreted by humans, so a human-centered approach is critical toward understanding the meaning of data as translated through models for mitigating potential biases and unforeseen consequences.

The six principles presented here are not a comprehensive articulation of all issues and solutions related to data excellence in AI but serve as an illustration of the emerging space and the concept of DCAI.

### Principle 1. Systematic Improvement of Data Fit

The use of data in DCAI is strategic, meaning data by itself does not provide value and is conceived as a means to address specific problems and objectives (primarily decided by the domain experts) and to improve the model (primarily decided by the AI experts). The ultimate value of data, therefore, lies in its coupling with the model and algorithm, and serving various stakeholders.[20] In this context, the idea of data fit refers to the extent to which the model is supported and comprehensively covered by the data, considering the distinctive attributes of the real-world environment in which the AI system operates. It also ensures the data offers a sensible representation of this environment.[33] For example, the data fit includes adequate coverage (for example, important variables to be covered) and the data balance (for example, evenly distributed number of samples from each class and absence of biases that can adversely affect the model's performance).

The major data practice that will improve data fit is a shift from defining and collecting (fresh, new) data toward augmenting data as an iterative progress. After each iteration, the DCAI team may examine if there is a need to clean and better prepare existing data; generate synthetic data via data augmentation; or collect fresh and new data.[29] A best practice when creating a dataset for the first time is to identify the sources of bias in the data. Random data acquisition may result in imbalances in the number and types of cases or events, creating un-representative samples or hard to replicate phenomena in the data. These hard examples are often considered a fundamental shortcoming of AI systems (that is, those only recognized by humans and not by the current trained model). For instance, a random collection of patient records in ICUs selected to detect secondary infection in hospitals results in a majority of uninfected patient cases and a minority of patients with secondary infection. Such data imbalances could introduce a false negative error bias in the model, predicting cases as not infected. Oversampling of the cases representing the phenomenon of interest to obtain sufficient numbers to result in an effective analysis may be more expensive and time-consuming. For this reason, identifying the potential minorities in the scoping phase of the AI project improves the quality of the dataset.

> **Data is the backbone of AI systems across the board, but the application of DCAI is even more vital beyond the frontiers of AI research and practice.**

A related problem is that some hard examples may not be clear during the initial data acquisition until revealed by error analysis on a trained model. Optimal model training requires observing additional data to better recognize hard examples. Data augmentation is a strategy that can diversify the data by oversampling the hard examples and creating a balanced dataset.[3] For instance, various image transformations (shear, distort, rotate, among others) and slicing in speech recognition have been commonly used as augmentation techniques to create synthetic hard examples.[30] Generative adversarial networks (GANs) are emerging as a useful tool to reveal and synthesize more hard examples.[15,28]

### Principle 2. Systematic Improvement of Data Consistency

As noted earlier, one of the defining metrics of data quality in DCAI has to do with the accuracy or consistency of data rather than its sheer volume (that is, big data). Accuracy refers to the degree to which the annotations are consistent with a gold standard;[21] in other words, how accurately the annotations reflect what they are supposed to reflect. As a closely related concept, data consistency refers to the adherence to the same procedure of annotation across annotators and over time.

Adoption of automated tools is now facilitating the consistency of annotation, a labor-intensive and time-consuming process. Weakly supervised labeling tools increase the pace and consistency of image annotation by providing pseudo labels. For example, in the case of instance segmentation, drawing a rectangle around the object is easier than drawing its exact contour with a polygon. Then, unsupervised algorithms such as Grabcut, can help produce the object contour using the weakly annotated rectangular bounding box.[34]

Even with the rise of automated tools, annotation remains a process that requires humans to be in the loop, which comes with inherent inter-annotator consistency issues. The concept of 'intercoder reliability' has long been used in data coding in academic research.[25] The same approach can be applied to ensure the consistency of training data wherein the annotation outcomes and their consistency are measured quantitatively. Measures of agreement such as kappa coefficients of correlation coefficients can be used as measures of agreement between multiple annotators.[21]

DCAI prioritizes implementing procedures to enhance data consistency. Conversations and agreements around consistency during each iteration will help align the whole team and will result in documenting and fine-tuning consistency standards and guidelines for those involved in data annotation, especially newcomers. These instructions can uncover edge, ambiguous, or borderline cases, and use them to clarify the conventions and good practices of annotation. As an example, DCAI teams can build on protocols such as Gated Instruction, that interactively teach annotation tasks, evaluate their inputs, and provide feedback on errors, ensuring the annotators have gained adequate competencies relative to the task at hand.[24]

Other relevant documentation practices here that have utility beyond inter-annotator consistency are systematically generating and recording metadata about datasets. Data is never raw.[4] For years researchers have emphasized the concepts of *data provenance* or *data lineage* as the process of tracking and recording the origins of data and how it has evolved over time.[9] More recently and in the context of AI systems, the 'datasheet for datasets' was developed to guide the documentation of key metadata about the dataset such as motivation behind its creation, collection process, distribution process, and recommended uses. This practice also facilitates communication among multiple stakeholders and potentially enhances the transparency of the AI system.[16]

### Principle 3. Mutual Improvement of Model and Data through Iteration

Recall the concept of iteration illustrated in Figure 2. Error analysis of the model after each iteration determines what subsets of the data are acceptable or unacceptable in relation to the model's performance, as such, helping ascertain the next iterations of data modification. For instance, error analysis can surface subsets of data that more significantly skew the model's performance, and hence, pinpoint data issues (for example, annotation inconsistency) that need to be overcome before retraining the model in the next iteration. When sources of bias are detected in the data, slices of data carrying those sets of bias can be reengineered to raise the performance of the model in relation to those issues. For instance, open source frameworks such as MLflow[11] and Data Version Control (DVC)[12] enable monitoring and maintaining ML models and their corresponding datasets over time.

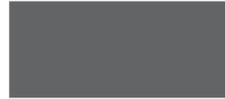

> Human-centered data science makes it clear that data should not be understood as objective and context-free.

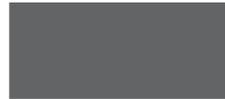

In DCAI, data collection is not done in one step (as in the model-centric AI approach), it is considered more targeted based on model-data performance considerations, and more importantly, continuously updated through many iterations integrating additional real-world examples. For instance, incorporating new instances of misinformation on social media or fraudulent financial transactions[33] results in the improvement of models beyond the initial production deployment of the AI system.

Machine Learning Model Operationalization Management (MLOps) is a critical set of emergent best practices in AI deployment focused on the continuous monitoring of deployed models and the feedback that is provided to update the model. Monitoring of issues such as *concept drift* or *data drift* is performed and the outcomes (for example, what the anticipated outcomes of the data may look like) are fed back into the next iteration to retrain the model and to refine the data.[29] Concept drift refers to changes in the relationship between input and output variables. An example of concept drift is changes in online shopping patterns due to unforeseen variables, such as, seasonality. Monitoring model performance over time after deployment is a strategy for detecting concept drift. Data drift refers to changes in the distribution of input data. An example of data drift is a facial-recognition algorithm that is trained on white faces and cannot effectively detect faces of other skin tones when the model is deployed. Continuous cluster analysis of data points combined with dimensionality reduction techniques (that is, PCA, t-SNE, UMAP) is an effective strategy for detecting data drift.[6]

A major outcome of DCAI iterations is therefore improvement in both model and data through mutual adjustment. Not only does the data improve, but the model adapts and boosts its performance by developing a better relationship between input data and output targets as well as better generalization over unseen data.

**Principle 4. Human-Centeredness of 'Data Work'**

The recent turn to data-centric approaches provides opportunities for a renewed attention to the representativeness of human contexts in the data that fuel AI systems.[39] Human-centered data science is an emerging area that critically analyzes the implicit but pervasive assumptions that "if it [data] is big enough, it will tell everything," and "the human element [can] be scrubbed from the database."[4] Instead, data is treated as a human responsibility and endeavor as humans are implicated in every step of data work and their choices have consequences for ethical and responsible uses of data in building AI models.

Data work refers to essential human activities that shape data, for example, acquiring, designing, curating, and improving data.[4] Human-centered research indicates that data work is often guided by a 'data vision,' which includes formalized computational methodologies and standardized formats (often affiliated with formal training) as well as discretionary craftspersonship and situated workarounds/fixes (often affiliated with real-world practice of data science) that adapt to the unique contingencies of the context of data at hand.[31] Because of the focus on the conventional work of optimizing models and algorithms, data work is often invisible and underappreciated. If DCAI approaches are to improve data quality (as a bottleneck of today's AI systems) and the human impacts of models, they need to move beyond "data as grunt work" and draw attention to data excellence through intentional practices, such as systematic data documentation, established feedback channels throughout the DCAI life cycle, and continuous conversations with domain experts and field partners.[35]

In addition, human-centered data science makes it clear that data should not be understood as objective and context-free. Data is inexorably embedded in a context and is not separable from humans who develop, use, and interpret it (making it relational and a matter of design activities and choices).[14] Therefore, data that feeds AI systems reflects important individual and social subjective (sometimes unrecorded) decisions, such as "What counts as the data?", "What is an outlier to be excluded from data?", and "What counts as the ground truth data?" A factor that can further complicate data work is that these questions may be handled based on unique political dynamics and dominant, sociotechnical value systems.[32] Biases such as the 'street light effect' can also pose clear limitations on how data scientists may choose and treat data sources. For instance, data scientists may use datasets that are easy to obtain and easy to convert to a relatively clean training set rather than asking good questions and clarifying what needs to be explored, what we called strategic use of data under principle 1.[4] To be mindful of these contextual and human-centered dimensions of data, DCAI teams need to take a critical perspective on "the data" and explore and capture different users' behaviors. Such a critical perspective could afford a deeper knowledge of the problem to be solved by the AI system while mitigating negative consequences of the system (such as, decision bias and discrimination).[4]

A critical perspective could also help detect and remedy data and model issues. ML engineers may often aim for collecting "clean" datasets which are free from outliers and so-called "noise" to improve the model performance.[39] This urge could result in "overfitting," which inevitably lowers the representativeness of data when the model is overfit by removing features that are present in the real-world applications. In addition to these types of biases that may stem from data sampling approaches, specific modeling decisions can also introduce bias. For example, choosing what parameters to keep, change, or discard can be a culprit in biased performances of the system. Preventing such problems requires not only computational and mathematical prowess but also a contextual knowledge of the domain in which features are being developed.[4] This brings us back to the importance of involving stakeholders who can better understand the context of use (see principle 6 for more information). To this end, tools that help communicate data and system performance to human stakeholders are of paramount importance (for example, data visualization and/or visual analytics).

**Principle 5. AI as a Sociotechnical System**

AI systems increasingly impact the lives of millions and are used in critical contexts including criminal justice, healthcare, human relations management (HR), and education. But algorithms are hardly neutral or objective and depend on the goals and discernment of the users. Human-centered AI presents these assemblages as complex sociotechnical systems that embody both technological elements and social norms, values, and preferences.[10,13] Similar to the DCAI approach, human-centered AI argues that development efforts must go beyond engineering optimization of AI models and embrace humans' needs, aiming to augment them (rather than replacing them) in ethically sensitive ways.[38,39] This requires specific design frameworks and strategies to ensure AI systems are ethically designed, meeting the needs of different stakeholders (particularly the marginalized and vulnerable). Examples of ethical considerations include respecting the people's right to privacy, providing fairly distributed benefits, and compliance with different legal frameworks regarding people's right to data.[4]

In human-centered AI the goal is not only to raise the automation efficiency (which may come at the expense of human agency in traditional frameworks) but also to augment human performance and promote their competency and responsibility. As a result, design frameworks must be directed at empowering humans to steer and control automated systems to exercise agency and creativity more effectively. This requires strategies that help both users and developers arrive at a mutual understanding about the goals and functions of AI systems. Examples are exercising audit trails and consistently soliciting granular feedback from the users about the state of the AI system.[38] In this way, a key mission of human-centered AI is opening the "blackbox of AI" and making AI systems explainable and understandable to humans,[13] for example helping the human operators understand what the AI system is capable of doing and why the AI system behaves in certain ways.[2]

In future DCAI teams, human-computer interaction (HCI) or user experience (UX) experts can play a critical role by integrating human-centered inputs and by taking stock of human behaviors and real needs. Human-centered design or human-centered data approaches (for example, end-user data probes or storyboards capturing human use contexts) provide much-needed perspectives for systematically incorporating the interest and needs of end-users as well as establishing common ground for collaboration between AI, HCI, and domain experts.[39]

### Principle 6. Continuous and Substantive Interactions between AI and Domain Experts

Data that feeds AI models is increasingly complex, multidimensional, and application/problem-dependent. As such, the most effective datasets fueling AI models are custom engineered through a partnership between AI engineers and domain experts[29] who understand the nuances of operations as well as the yardsticks of quality within their own unique context. For example, in medical contexts, patient data is often hard to come by given the privacy and compliance frameworks. As a result, stringent regulatory protocols (for example, HIPAA Patient Rights in the U.S.) are attached to data practices and data governance in these domains. This requires the continuous involvement of domain experts engaging in pragmatic data collection and evaluation or even assessment of whether the model achieves what it is expected to achieve. The domain experts enhance the evaluation process by developing specific use cases that put the model into more domain-sensitive tests.[37] They contribute not only to data collection and preparation processes but also the explanation of the resulting AI model.[27]

Taking data seriously means taking data annotators and relationships with them more seriously. In contrast, recent research indicates that many AI developers see annotators as "non-essential" and do not necessarily value their expertise and contributions.[36] Bernstein suggests a shift in mindset from 'annotation as a service,' which implies transient relationships with annotators, toward 'annotators as collaborators,' which requires a hands-on approach toward data training and a longer and more substantial relationship with the data annotators.[7] To this end, a crucial way to achieve data consistency in model training is a clear conversation between AI experts and individuals tasked with annotating data. For example, multiple and conflicting labeling conventions, particularly in the case of complex and tacit tasks, can confound the data quality and consequently risk lowering the performance of AI models. AI experts must clearly and continuously communicate their intent to domain experts and data annotators and strive to provide them with incremental feedback in each iteration.[7]

### Future Directions and Steps

Some of the future directions that can advance DCAI and help address the current gaps in AI practices include the following:

**Data preparation and augmentation.** Data preparation and quality control should become a central concern of AI teams rather than an afterthought. This requires building upon best practices already formed around concepts such as data provenance and data lineage. Moving forward, we need to create a more rigorous metadata development regime. A significant obstacle to enhancing data quality is the lack of available information on the provenance of the data and the associated models. Unfortunately, many current studies do not provide this information, nor the data or code necessary for confirming results and making comparisons.

Possible solutions in the future will likely involve a partnership between human experts and automated systems in cleaning and preparing data. For example, another open source project, AutoML already provides opportunities for automating building ML models. Future efforts can revolve around adopting AutoML, or a similar approach, to help DCAI teams with concerns around label noise or coverage of datasets as well as continuous data collection or label correction.[33] This noted, humans must stay in the loop; manual inspection of data will still constitute a crucial task for AI experts.[37]

**MLOps teams and tools as integrative solutions to data quality.** DCAI makes it clear that data defines the behavior of deep learning systems, and that noisy or messy data are key culprits thwarting AI performance. As such, a core mission of MLOp teams, tasked with effective deployment and maintenance of AI models in production, must be the processing, handling, and monitoring of data that train systems before and after they go live.[37] As a result, we foresee that MLOps teams and practices will be involved in the whole life cycle of DCAI; and thus, the scope and responsibility of these teams will expand as AI systems continue to be used in real-world application, beyond isolated experimental situations. Here established standards and principles of excellence in software engineering can provide inspiration and analogous metrics (for example, maintainability and fidelity) for developing frameworks of data quality assessment.[5]

**Data bias and audit procedures.** AI decision-making is not neutral. AI systems can reflect, reinforce, or amplify social biases. In recent years, serious concerns have been raised about bias in AI due to undetected or unaddressed sources of bias in the dataset. For example, researchers found that African Americans were overrepresented in mugshot databases, making them more likely to be singled out by algorithms used by law enforcement.[22]

Given the focus of DCAI on data, it is crucial to address challenges of data bias through technical solutions such as counterfactual testing, or through operational practices such as value-sensitive design methods or third-party audits of AI models. In collaboration with other communities invested in the topic, DCAI researchers and practitioners must develop tools and policies that help detect why certain inconsistencies may be introduced in datasets, and why certain attributes in the data might have been over- or under-represented in the system.

**A human-centered mindset.** As AI systems are increasingly deployed in high-stakes decision making that carry real-world consequences, these systems should incorporate continuous interactions with users and contextual factors to develop more ethical, responsible, and equitable outcomes. Examples such as Google Flu Trend raise issues of 'algorithmic hubris' where the developers' vision of fully autonomous, end-to-end systems glosses over the dynamic nature of user behavior and wider social contexts.[38] Google Flu Trend system was shut down after two years because it raised concerns over the meaning of its predictions and how precisely it reflected the real flu-related patterns and behaviors.

A healthy skepticism toward unexamined computation and emphasizing human-centered perspectives toward AI systems can help address the ethical concerns raised by AI systems described in this article. This can be done by bringing to focus people and contextual factors that may play consequential roles in shaping, evaluating and integrating data and by advocating a systematic approach to identifying and often conflicting values and contexts. These approaches call for crucial changes in attitudes and system development, deployment, maintenance practices in order to give voice to multiple stakeholders including domain experts, end users, and those that are affected by the system. Recent work in human-centered AI invites moving beyond the reductive metaphor of "user" and directly engaging with questions such as "who is the human in human-centered machine learning."[10] Inspiration can be drawn from frameworks that reveal and incorporate multiple stakeholders' values (for example, value-sensitive design) to more effectively investigate design trade-offs that help empower all stakeholders.

Finally, AI-based systems present new challenges in usability testing and interface design. For example, because of their unpredictable and ever-evolving nature, these systems may challenge the principle of consistency that encourages interface stability over time. Therefore principles, guidelines, and strategies for designing human users' interactions need to be revisited to accommodate users' interactions with these emerging systems.[2]

**Crowdsourced data supply chain.**

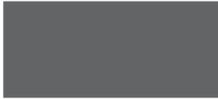

A crucial way to achieve data consistency in model training is a clear conversation between AI experts and individuals tasked with annotating data.

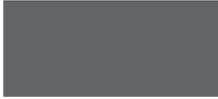

Data annotation now greatly relies on crowd workers. Over the past few years, this has spurred a rise in industry-based data annotation and labeling services (for example, Amazon Sagemaker Ground Truth, and Scale.ai). Some current practices raise both ethical and practical concerns. Recent research demonstrates that the divergent interpretations of annotators can result in inconsistency in the data annotation, adversely impacting AI performance.[17] Clear communications between ML engineers and crowd workers, creating "the codebook," or performing some of the annotations by ML engineers themselves can be important initial steps in enhancing the consistency of data.

Another major way to improve the data supply chain is by making systemic improvements to the experiences of crowdworkers that annotate data. Ample research in recent years suggests that unfair treatment of these workers (for example, lack of access to the minimum wage) together with issues such as tight algorithmic control exercised on microtasking platforms could create a precarious work situation, which in turn impact the reliability of the whole data supply chain (beyond clear and valid ethical concerns).[19] DCAI efforts and projects need to formulate strategies to improve the life of these "ghost workers" behind the 'AI curtain' by providing fair compensation and a more meaningful work experience.[18]

**Conclusion**

As Aroyo et al.[5] succinctly points out, "data is potentially the most undervalued and de-glamorized aspect of today's AI ecosystem." A lack of attention to the human-centered nature of data can result in important negative consequences in the outcomes of AI systems such as bias in performance. DCAI can bring to the forefront the role of data in the performance of AI systems and gives precedence to the quality of data over its size. Data should be viewed as human-centric, dynamic and always evolving rather than a static input to the model to be engaged with only in the preprocessing stage.

Improving data remains a crucial iterative undertaking, going beyond the "preprocessing" mindset and instead throughout the system life cycle. In this article, we present DCAI as a budding movement and bring together already existing but dispersed data concepts and practices in the form of six principles to guide DCAI efforts. The future of DCAI lies in creating systematic and sociotechnical processes for monitoring and improving data quality, which requires closer attention to how data is supplied, prepared, annotated, and integrated into AI systems.  c

AI decision-making is not neutral. AI systems can reflect, reinforce, or amplify social biases.

**Mohammad Hossein Jarrahi** is an associate professor in the School of Information and Library Science (SILS) at the University of North Carolina, Chapel Hill, NC, USA.

**Ali Memariani** is a computer vision scientist at Syngenta, Raleigh, NC, USA.

**Shion Guha** is an assistant professor in the Faculty of Information and Department of Computer Science at the University of Toronto, Canada.